\documentclass[runningheads]{llncs}
\usepackage{graphicx}
\usepackage{verbatim}

\begin{document}
\title{Strategy to Increase the Safety of a DNN-based Perception for HAD Systems}
%
%
\author{Timo S\"amann\inst{1} \and 
Peter Schlicht\inst{2} \and 
Fabian H\"uger\inst{2} }
\authorrunning{T. S\"amann et al.}
%
\institute{Valeo Schalter und Sensoren GmbH, Kronach Germany \and
Volkswagen AG, Wolfsburg, Germany\\
\email{timo.saemann@valeo.com}\\
\email{\{peter.schlicht,fabian.hueger\}@volkswagen.de}}
\maketitle              
\begin{abstract}
Safety is one of the most important development goals for highly automated driving (HAD) systems. This applies in particular to the perception function driven by Deep Neural Networks (DNNs). For these, large parts of the traditional safety processes and requirements are not fully applicable or sufficient.
The aim of this paper is to present a framework for the description and mitigation of DNN insufficiencies and the derivation of relevant safety mechanisms to increase the safety of DNNs. 
To assess the effectiveness of these safety mechanisms, we present a categorization scheme for evaluation metrics.

\end{abstract}
\section{Introduction and Big Picture}
A massive number of academic publications deal with the use of DNNs for the realisation of HAD. The strong performance of DNNs on perception tasks raise optimism that HAD systems will become reality in the foreseeable future. 
In order to see HAD vehicles on the road, there is relevant challenges with regard to certifications, public trust, acceptance, validation and legal regulations to be overcome. One of the major challenges in the industrialization of safety-critical autonomous systems such as HAD is safety.

\begin{figure}[h]
    \includegraphics[width=0.5\textwidth]{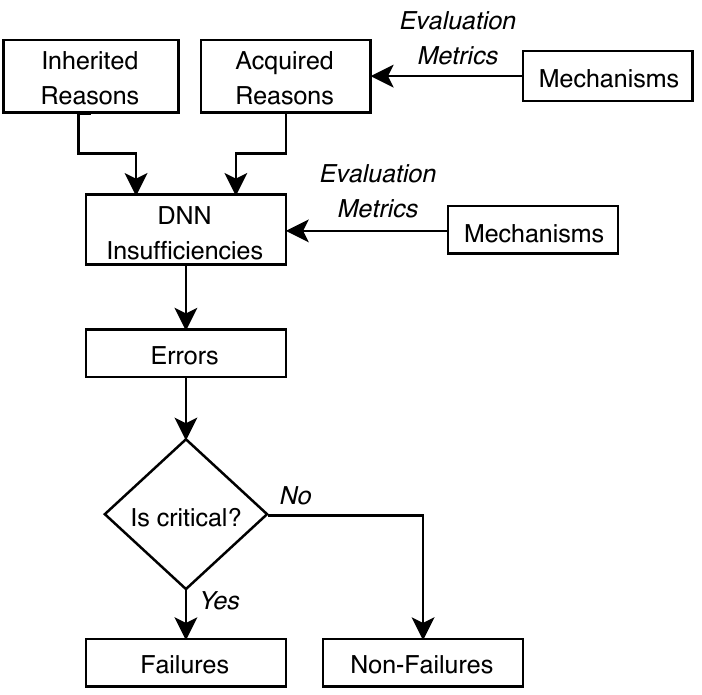}
    \centering
    \caption{Inherited and acquired reasons causes the DNN insufficiencies that lead to errors and depending whether they are critical to failures. The acquired reasons and DNN insufficiencies are mitigated by safety mechanisms that automatically lead to a mitigation of errors. The success of the mitigation is measured by evaluation metrics.}
    \label{fig:intro_image}
\end{figure}

There are numerous standardized quality management and engineering process requirements that ensure an acceptable level of safety for road vehicles.
So far, it has been largely neglected that HAD systems require a vehicle environment perception and decision making that is carried out with an intended functionality. 
At the beginning of 2019, the ISO/TC 22 Technical Committee presented a draft on the absence of unreasonable risk due performance limitations of the intended functionality, which is specifically free from the faults addressed in the ISO 26262 series - called ISO/PAS 21448 (abbreviation~SOTIF)~\cite{SOTIF}. 
The performance limitations include sensor performance-, actuator technology- and algorithms-limitation (e.g.~machine learning). However, the consideration of machine learning algorithms is restricted to the automation level SAE-2, where the driver serves as the fallback-level and is thus responsible for the final monitoring of the system (e.g. failure detection and risk mitigation)~\cite{czarnecki2018road}. In addition, the considerations are very general and lack technical depth and practical use.
For these reasons, it would be beneficial to start new standardization efforts or to extend existing ones.

Argumentation approaches in an assurance case~\cite{AS} are promising concepts to approach this challenge~\cite{burton2017making}.
The aim of an assurance case in this context is to provide a convincing argumentation about the fulfilment of the safety goals. The argumentation includes evidence of compliance with the safety requirements derived from the safety goals, taking into account the operational context. The evidence can be carried out in a statistical manner and by means of well-documented design processes.

A considerable part of this argumentation deals with the mitigation of errors of the DNN predictions. Those errors may be purely performance-related (e.g. false-positive or false-negative detections or localization inaccuracies) or less obvious flaws such as unreliable confidence behavior or a non-answer of the DNN. In other words, it is an erroneous behavior that might lead to an overall system failure (see section~\ref{sec:Errors_and_Failures}).

Errors are caused by DNN insufficiencies\footnote{We consider DNN insufficiencies and negative DNN traits as synonyms.} which are described in more detail in section~\ref{sec:Functional_Insufficiencies}. 
These describe the negative traits that DNNs naturally have and which the training process could reinforce.
We divide them into five categories, which cover the \textit{input}, the \textit{internal functioning} and the \textit{output} of the DNN and provide one example for each category:

\begin{enumerate}
\item Lack of generalization:
The lack of generalization ability becomes apparent when the input data differs in its data distribution from its training data set. The data distribution can be inside or outside the input space defined in the ODD. 

\item Lack of robustness: A lack of robustness means a reduction in performance due to perturbations that are present in the \textit{input} data, such as noise on the sensors (e.g. dust, dirt or smashed insects on camera lens).

\item Lack of explainability: The lack of explainability refers to the \textit{internal functioning} of the DNN. For example, it is unclear which rules were decisive for the DNN to reach its outcome.

\item Lack of plausibility: The lack of plausibility refers to the \textit{output} of the DNN: the \textit{prediction}. For example, plausibility concerns the prediction of physically impossible pedestrian sizes or shapes.

\item Lack of efficiency: The lack of efficiency refers to the expected inference time, the number of required operations and parameters as well as the required memory. This is especially relevant considering the benefits of redundancies to improve safety and the limited hardware resources available in the application.
\end{enumerate}

The causes of the insufficiencies are divided into two categories following the nature vs nurture debate~\cite{ceci1999nature}: Inherited and acquired reasons (see Fig.~\ref{fig:intro_image}).

Inherited reasons arise from the underlying principles of current AI technology (see subsection~\ref{sec:Inherited_Reasons} for more details).
Acquired reasons (see subsection~\ref{sec:Acquired_Reasons}) are related to the development process of DNNs and have an impact on DNN insufficiencies: Operational Design Domain (abbr.~ODD) and data set specification, DNN topology design choices, training process and postprocessing. For example, an unbalanced data set with respect to the applied optimization goal might lead to a reinforcement of DNN insufficiencies. 
It is therefore important not only to mitigate the negative traits that the DNN innately has, but also to positively influence the acquired reasons.

To minimize the DNN insufficiencies, so-called safety mechanisms are used. Safety mechanisms include methods for the analysis of safety-relevant aspects of the DNN, as well as safety-enhancing steps based on the analysis results. The relevant analyses cover all stages of the development process.

In order to make the increase in safety measurable, corresponding evaluation metrics must be defined. They give the safety mechanisms a benchmark and possible plausibility check of their success in increasing safety. 
In addition, the evaluation metrics are used by the assurance case to assign statistical evidence or guarantees to the assumptions. 

The main contribution of the paper is to formulate a strategy on how DNN-based perception functions for HAD can be made safe. 
At the same time the paper should not be seen as comprehensive and valid strategy to increase DNN safety that can be used in industry. 

\section{Related Work}
The research on automotive safety standards and processes for DNNs is quite scarce as it lies in between two research fields: Safety and machine learning. There are currently few publications available that attempt to bridge the gap between these areas:\par
Within the UL4600 (current draft version available from~\cite{UL4600}) and the recent whitepaper ``Safety First for Automated Driving'' contain collections, challenges and recommended design principles to reach safety for machine learning. Even though they contain a lot of useful guidelines and requirements, they lack a concise overview of the sources of DNN insufficiencies.
Existing literature within the research field of machine learning seems to be more concentrated on subdomains such as \textit{explainable AI}, \textit{adversarial robustness} or \textit{dependable AI}. Here, a concise overview and, in particular, research on interconnections between these topics seems to be missing.

\section{DNN Insufficiencies}
\label{sec:Functional_Insufficiencies}
DNN insufficiencies describe systematic and latent weaknesses of DNNs that they possess due to inherited and acquired reasons. 
They can cause an error in the inference when faced with triggering data (triggering event).
It is important to identify the DNN insufficiencies in order to deal with them adequately in the argumentation in the assurance case using safety mechanisms as mitigation methods. 
We propose a division into five categories for which we claim to provide comprehensive coverage of DNN insufficiencies. The properties become apparent when the input data is changed (lack of generalization ability and robustness), by questioning the inner workings (lack of explainability), by analyzing the prediction (lack of plausibility) and by considering the hardware requirements in the application (lack of efficiency). 
We provide some examples each to increase the understanding of what is meant by the individual points:

The \textit{lack of generalization} is noticeable when the input data changes in its underlying data distribution compared to the training data.
This leads to a reduction in the accuracy of the DNN prediction, depending on the quantity of distribution change. The data distribution can be inside or outside the input space defined in the ODD.
Changes in the visual appearance of objects or traffic scenes in general can lead to a shift in the data distribution that is still in the domain. These can be caused by different weather and lighting conditions or by different car brands, wall decorations and pedestrian clothing. Other appearances can lead to leaving the domain and cause a domain shift. For example, given a DNN-based scene understanding, which was trained with European-looking cities would have a lower accuracy in the Asian region. Furthermore, a change in the sensor setup, e.g. a change in the viewing angle due to a change in the mounting position or a change in the sensor characteristics, e.g. a different camera lens or sampling rate of the LiDAR sensor, also have a negative impact on the accuracy.

The \textit{lack of robustness} means a reduction in performance of the DNN due to perturbations that are present in the input data. Such perturbations can be noise on the sensors caused by dust, dirt, rain drops or smashed insects. Further perturbations may occur due to errors in the transmission of sensor data to the DNN (e.g. packaging loss due to buffer overflow). A serious perturbation is adversarial attacks, which are hardly visible or non-suspicious to humans, but can completely change the DNN prediction. The corresponding patterns can be applied as foil or stickers to objects~\cite{eykholt2018robust} or directly to the camera lens~\cite{li2019adversarial}.

The \textit{lack of explainability} refers to the internal functioning of the DNN and can become apparent by questioning the inner workings. It is not possible to rigorously explain which features in the input were relevant for the DNNs decision and which rules led to the respective decision. Also the verification of the learned representation (e.g. for object recognition, what does a prototype object for the DNN look like?) is not possible innately.

The \textit{lack of plausibility} refers to the \textit{output} of the DNN: The prediction (e.g. semantic segmentation) and the confidence values which can be obtained by probabilities from softmax distributions. For example, a DNN for pedestrian recognition detects pedestrians at positions that are excluded by physical laws. Another examples are predictions of shapes and sizes of objects that are not plausible. The problems with confidence estimation are that false predictions are sometimes predicted with high confidence (keyword: unknown unknowns) and generally the modern DNNs tend to be overconfident~\cite{guo2017calibration}.

The \textit{lack of efficiency} refers to the DNN itself and describes the efficiency properties of the DNN. These are for example inference time, the number of required operations and parameters as well as the memory requirements. The consequences of a lack of efficiency result from the consideration of hardware resources in the application. Hardware resources are very limited with regard to HAD systems that realize a large number of functions. This can lead to restrictions in the general DNN selection as well as in the redundancy strategy, which in turn affects safety.

Generally, insufficiencies are hard to measure and detect as they only get apparent on particular data (triggering event) when they cause measurable errors.
The reasons for the DNN insufficiencies can be divided into two reasons in analogy to the age-old nature vs. nurture debate: Inherited and acquired reasons.

\subsection{Inherited Reasons (Nature)}
\label{sec:Inherited_Reasons}
Inherited reasons concern the underlying principles by which the current AI technology works. It is questionable whether these principles can be used to simulate e.g. common sense, which would be helpful for a fundamental improvement of DNN insufficiencies. Current work in this area falls under the field of Artificial General Intelligence (abbr.~AGI) and is often based on the foundations of the biological model - the human brain~\cite{eth2017technological}.
The directions of these works are very different and short-term successes are estimated to be low~\cite{grace2018will}. For these reasons we do not treat the inherited reasons and accept them as given.

\subsection{Acquired Reasons (Nurture)}
\label{sec:Acquired_Reasons}
Acquired reasons describe all the influences we can exert, based on the underlying AI technologies. 
We refer to the following processes used in the development of DNNs:
Specification- (ODD, data), DNN topology design choices, training- and postprocessing process.
Errors within these processes reinforce the DNN insufficiencies. 
We consider DNN topology design choices to be acquired reasons and not to be inherited reasons because we claim that all DNN topologies are based on the same underlying principles.

\section{Errors and Failures}
\label{sec:Errors_and_Failures}
When a DNN suffers from a DNN insufficiency and is provided with particular data (``triggering event''), it produces an \textit{error}, i.e. a measurable malfunction of DNNs. They can be manifold ranging from imprecisions to incorrect self-assessment and usually stem from a DNN insufficiency being exploited by particular data. Examples of \textit{error} types are the following: False positive detection; False negative detection; Unreliable confidence estimate; Too long computation time; Wrong size or location estimate.

Depending on which system the DNN is integrated into and how critical the situation is, this will lead to a \textit{failure} of the affected system 
A \textit{failure} refers to a malfunctional state of the overall HAD system (e.g. unnecessary braking of an autonomous vehicle). It is important to understand that not every \textit{error} leads to a \textit{failure}. It is an important goal of safety mechanisms on module level to mitigate DNN insufficiencies while safety mechanisms on system level aim at mitigating the effect of module \textit{errors} in order to prevent \textit{failures}.

\begin{figure*}[h]
    \includegraphics[width=1\textwidth]{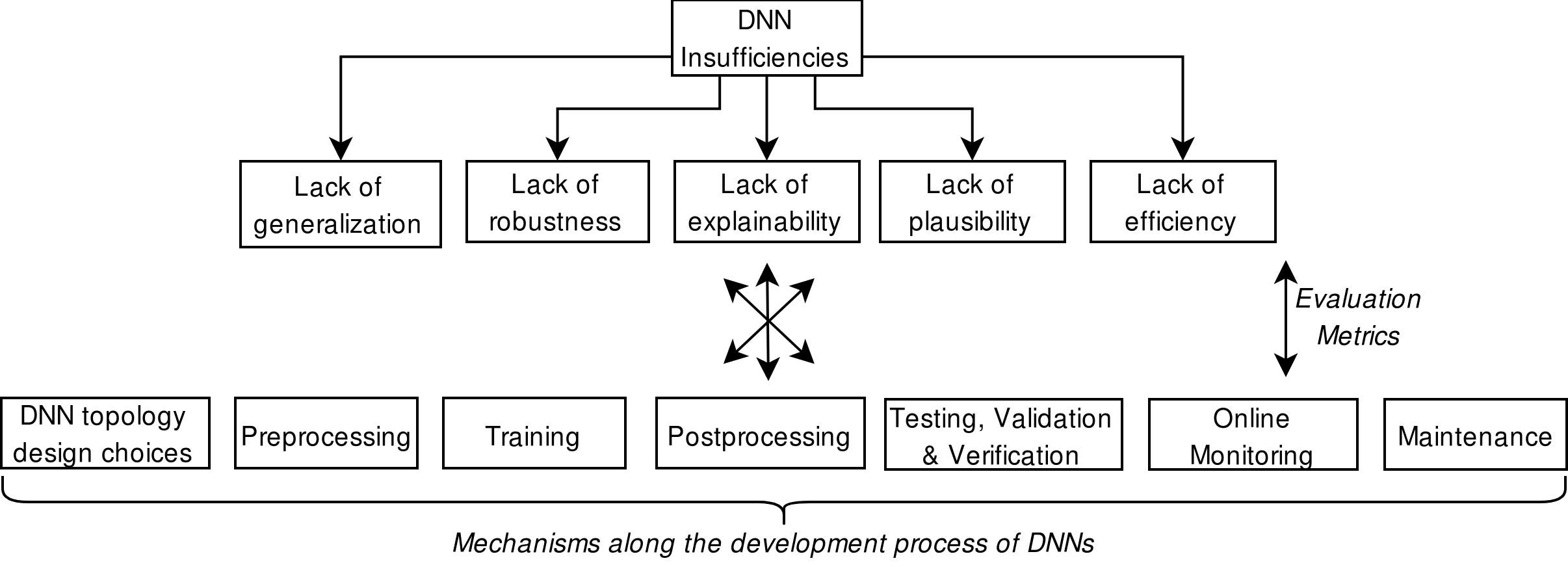}
    \centering
    \caption{The DNN insufficiencies can be divided into five categories. DNN insufficiencies are minimized by safety mechanisms that are structured along the development process of DNNs. The concrete connection between them (see the three arrows) remains part of future work. The evaluation of the safety mechanisms on the basis of their success in mitigating DNN insufficiencies is achieved via evaluation metrics.}
    \label{fig:overview}
\end{figure*}

\section{Safety Mechanisms}
\label{sec:Mechanismen}

In order to mitigate DNN insufficiencies, safety mechanisms (in short: mechanisms) are used (see Fig.~\ref{fig:overview}). They achieve this on the one hand by mitigating the acquired reasons and on the other by directly improving the DNN traits. All actions that pursue this goal are called mechanisms. This includes, for example, analysis tools that measure safety-relevant aspects in the data set or the DNN itself and, based on their results, perform out safety-enhancing measures. In addition, these can be network structure and training process changes. A detailed list of mechanisms along the development process of DNNs is presented in the following. We also provide a brief description of why the mechanisms can be used to minimise DNN insufficiencies as well as some hints on concrete usable approaches.

\subsection{DNN Topology Design Choices}
The choice of the DNN topology design is divided into the architecture selection for the DNN and the redundancy strategy that can be followed to make the solution of the task even safer.
Architecture design:

\textit{Architecture Search:} The architecture design is based on an architecture search strategy whose goal is finding the optimal network topology by means of an automatic process. It has been shown that architecture search can be superior to human handcrafted design~\cite{chen2018searching}. 
If the criteria by which the automatic process searches for a suitable design are met, not only on prediction quality metrics (see subsection~\ref{sec:Prediction_Quality_Metrics}) but also on safety metrics (see subsection~\ref{sec:Safety_Metrics}), this may reveal previously unknown designs.
\textit{DNN class:} The choice of the DNN class has an influence on important properties of the DNN. Most DNNs for classification and regression tasks are CNN based. However, CNNs do not take advantage of the temporal consistency present in video data. The choice of recursive networks or recursive architectures such as LSTMs, GRU and their bidirectional variants can bring this advantage. Capsule networks enable equivariance instead of translations invariance~\cite{sabour2017dynamic}. A careful selection must be made according to the specific task and security requirements.
Redundancy strategy: 

\textit{Ensemble:} An ensemble of networks consists of different architectures that are based on different modalities or have been trained with different hyperparameters or data sets. By combining the predictions in a suitable way, the different properties can be combined for improvement~\cite{hansen1990neural}. Networks with a different accuracy vs. runtime tradeoff are also conceivable, so that networks with particularly good quality and great robustness can serve as a reference for faster networks and thus enable self-monitoring. 
At the same time, this offers an increase in functional safety, since in the event of a network malfunction, other networks can perform the same task.
\textit{Sensor fusion:} The sensor fusion takes place on different levels depending on the fusion concept. A sensor fusion at sensor level is just as conceivable as a fusion at feature or regression level. LiDAR and camera sensors are often used for fusion~\cite{gao2018object}. The fusion concepts provide the network a richer feature base, allowing the DNN to extract better feature correlations from the higher quality database. In addition, the advantages of the corresponding sensor technologies can be combined.

\subsection{Preprocessing}
Preprocessing refers to the optimization of the training data set.

\textit{Adversarial training:} Adversarial examples are input data that have been manipulated in such a way that they fool the DNN and are not visible to us humans. In adversarial training the adversarial examples are included in the training data. According to the number of adversarial examples added to the training and the technique by which they are generated, an increased robustness of the DNN is reported~\cite{ganin2016domain}\cite{tramer2017ensemble}.

\textit{Data Augmentation:} Data augmentation extends the training data to achieve greater quantity and diversity. The most common augmentation techniques are cropping, padding, horizontal flipping and rotations. More advanced techniques can further enhance the benefits of data augmentation. For example,~\cite{cubuk2018autoaugment} uses 16 geometric and color-based transformations, using reinforcement learning as the search algorithm to determine them. \cite{rakin2018parametric} have presented a parameterizable gaussian noise injection technique where the noise intensity can be learned. The added noise can be seen as a kind of regularization, making the DNN more robust against adversarial examples.

\subsection{Training}

Training refers to the training process of the DNN.

\textit{Loss function:} The optimization goal in the training process for DNNs is called the \textit{loss function}. Its selection requires careful attention to describe important properties of the task to be solved as well as additional regularization objectives that may arise from safety considerations. Most existing loss functions such as mean square error or cross-entropy were designed with the primary goal of reaching fast convergence to highly accurate models - additional safety requirements such as robustness or explainability are not considered in this respect. 
An example of more safety-focused training paradigms would be the increased recognition of pedestrians along the driving path. The weighting of the individual classes is another possible variable to improve the safety of the resulting network.
For mere accuracy, a balanced class distribution~\cite{khan2017cost} provides the best benchmark values, but in terms of safety this is not necessarily better~\cite{chan2019ethical}.

\textit{Student-teacher approaches:} The idea behind the Student Teacher approaches is that an already trained DNN with high quality (teacher network) is taught to a DNN with lower quality (student network) by using the predictions of the teacher network as pseudo ground truth for the student network. It is expected that this will lead to a compression of the knowledge of the teacher network into the student network. This process does not require any labelled data, so large amounts of data can be used for training. Extensions of this approach are used, for example, to increase the robustness of DNNs against adversarial examples~\cite{bar2019robustness}. 

\subsection{Postprocessing}
Postprocessing refers to the process after training the DNN.

\textit{Calibration:} Modern networks tend to be over-confident. This sometimes leads to false predictions which are still assigned a high confidence. Calibration techniques, as compared in~\cite{guo2017calibration}, can help. Even simple methods such as temperature scaling - a single parameter variant of Platt Scaling - can be seen as an effective calibration.

\subsection{Testing, Validation and Verification}
The following mechanisms are used to test, validate and verify the trained DNN in an offline mode.

\textit{Heatmap methods:} Previous uses of heatmaps have concentrated on classification tasks, where the heatmap displays the image areas that have played a relevant role in classifying the image into a certain class~\cite{chattopadhay2018grad}\cite{montavon2017explaining}. This information can be used for offline verification of the learned. The evaluation of these heat maps by a human expert can reveal errors in the model (see~\cite{lapuschkin2016analyzing}). For example, in the case of pedestrian recognition, it could be revealed that the DNN considers only parts of the pedestrian to be relevant for decision making, which would result in the pedestrian not being recognized in the case of corresponding occlusion.

\textit{Formal verification:}
The verification of safety-relevant properties of a DNN as well as the description of the input space (naturally occurring and unusual inputs) for which the DNN shows a safe behaviour is the task of formal verification. Formal and rigorous guarantees can be issued by exploiting the DNN piecewise linear structure and taking insights from formal methods such as Satisfiability Modulo Theory (abbr.~SMT)~\cite{katz2017reluplex}\cite{bunel2018unified}. Previous approaches mainly deal with classification and regression tasks and make a variety of assumptions that do not apply in reality and thus do not allow for complete guarantees~\cite{huang2017safety}. A decisive limitation of formal verification is the scalability of the approaches, so that they are not transferable to state-of-the-art DNNs so far.

\textit{Testing:}
In addition to existing testing methods in the field of machine learning (i.e. careful withholding of an unseen subset of training data for assessing the resulting performance of DNNs) and automotive products (i.e. covering a maximum complete range of semantic scenarios) testing in an ``adversarial fashion'' may support the resulting safety of the DNN. Hereby, testing is performed in order to find particular weaknesses of DNNs. These \textit{corner cases} can stem from semantically complicated situations or slight perturbations of the input that confuse the DNN.

\subsection{Online Monitoring}
Online monitoring refers to the monitoring of DNNs during runtime.

\textit{Uncertainty modelling:} HAD systems will operate in an open world containing uncertainties due to the limited availability of information and the influence of random effects. In addition to this, model uncertainties arise due to imperfections within the training data set. Monitoring and reporting of resulting uncertainties to the perception of HAD systems addresses this circumstance as the softmax output of DNNs tends to overestimate the model confidence~\cite{guo2017calibration}. Possible approaches to report uncertainties besides the perception output are manifold: Bayesian networks output a statistical uncertainty, sampling-based technologies approximate this approach~\cite{gal} but come with run time challenges. An addition to this, sample free improvements of uncertainty estimates have been suggested~\cite{metaseg}\cite{samplefree}.

\textit{Out-of-Distribution (abbr.~OOD) detection:} If the input data is drawn from a different distribution than the training data, this can lead to a reduction in DNN performance. The extent of this reduction depends on the size of the shift in the data distribution. Under these circumstances, safe operation can no longer be ensured, which is why the recognition of OOD situations during runtime and application of the DNN is needed. If ODD situations are detected reliably at the input this information can be taken into account in a subsequent fusion in order to adjust the vehicle behavior (e.g. reduction of speed). OOD data can be recognized by PCA, various types of (variational) autoencoders and generative adversarial networks~\cite{kiran2018overview}. 
\subsection{Maintenance}
This subsection deals with the maintenance of DNNs which become necessary during use or after a certain time in use.

\textit{Sensor setup changes:} Due to the increasing demands from the automotive industry, sensor development is achieving a high level of dynamics. This results in regular innovations of the sensors that are installed in the vehicles. The modified sensor properties, however, lead to a reduction in the performance for the DNNs that were learned with the previous sensor technology. Re-learning the DNNs requires the creation of new training data generated with the modified sensor technology. This procedure is extremely costly and time-consuming. A methodical procedure for transferring the learned DNNs can be found in Transfer Learning. By specific modification of the parameters of the DNNs, an adaptation or abstraction to the new sensor technology is achieved. Possible approaches include self-ensembling methods and additional loss functions: Self-Ensembling methods, for example a student-teacher approach are used to modify the DNN parameters in a semi-supervised fashion~\cite{french2017self} or a GAN-based data augmentation takes place to fine-tune the DNN~\cite{choi2019self}. Additional loss functions minimize the difference between source and target correlations in the intermediate feature maps of the DNN. Conventional approaches are for example the Coral loss~\cite{sun2016deep} and Maximum Mean Discrepancy based Loss functions~\cite{long2015learning}.
\textit{New appearances of objects:} It is foreseeable that objects will change their appearance or that new objects will be come along. The DNN must be modifiable so that it can deal with these changes. In concrete terms, this can mean that a DNN is learning a new class, which must not have any negative effects on the other classes. No recommendations for possible mechanisms can be made on this point.

\textit{Failure report:} Further open questions concern the reporting of failures as well as the detection of failures in the application. Furthermore, the determination of the cause of these failures plays a decisive role in proposing suitable mechanisms on the basis of this information.

\section{Evaluation Metrics}
\label{sec:Indicators}
Evaluation metrics are used to assess the mechanisms of effectiveness in mitigating DNN insufficiencies. Evaluation metrics also offer the possibility of benchmarking the mechanisms, which compares the mechanisms and allows the selection of the most effective ones. 
We distinguish between metrics that refer to the quality of the DNN prediction (prediction quality metrics), the safety of the DNN (safety metrics), the efficiency of the DNN (efficiency metrics) and the quality of the data set (data metrics).
The differentiation of the metrics are mainly based on the input required for the calculation and on the objective they are intended to achieve.

\subsection{Prediction Quality Metrics}
\label{sec:Prediction_Quality_Metrics}
The input data for calculating the prediction quality metrics (abbr.~PQM) are the predictions of the DNNs such as semantic segmentation, bounding box detection or depth estimation and the ground truth. Appropriate PQM are for example IoU, area under the ROC curve, root mean squared error, F1 score etc.
PQM can be used to test the robustness and generalization capability of the DNN by modifying the test data accordingly (e.g. overlaying the data with perturbations).

\subsection{Efficiency Metrics}
\label{sec:Efficiency_Metrics}
The efficiency metrics indicate the efficiency of the DNN together with its mechanisms. A distinction is made whether the efficiency only concerns the DNN itself or is measured together with the mechanisms during validation or online monitoring. The efficiency includes the runtime, the number of required GFLOPs and parameters as well as the model size required for saving.

\subsection{Safety Metrics}
\label{sec:Safety_Metrics}
Safety metrics should be seen as an important complement to the PQM and allow statistically valid statements about key safety aspects of DNNs. They differ from PQM by the fact that their calculation is performed on the basis of mechanisms that rely on the internal structure of the DNN, e.g. heat maps, interpretations of feature maps, uncertainty maps etc. 
Safety metrics are hardly researched in the current research landscape, especially not to the extent and quality that is required in the automotive industry.

\subsection{Data Metrics}
\label{sec:Data_Metrics}
Data Metrics evaluate the quality of the data sets that are used to train, validate and test the DNN. The data sets are derived from the input space specified in the ODD. A balanced training data set is crucial for the DNNs appropriate robustness and generalization capability within the specified input space. The measurement of the coverage level and the density of the coverage of the input space are conceivable data metrics for which, to the authors' best knowledge, there are no solid approaches. Other conceivable data metrics concern the creation of ground truth in terms of accuracy and consistency. When using synthetic data, data metrics can evaluate realism. 

\section{Conclusion and Future Work}

We have developed a strategy to increase the safety of a DNN-based perception for HAD systems. This strategy provides for mitigating the DNN insufficiencies, which we have divided into five categories and described, which we equate with increasing safety. The reasons for the DNN insufficiencies have been divided into inherited and acquired reasons in analogy to the nature vs nurture debate. To mitigate the DNN insufficiencies we have presented a variety of potential safety mechanisms along the development process of DNNs. 
Furthermore, we have introduced a notation of potential evaluation metrics that can be used to measure the safety of DNNs. This allows the evaluation of the safety mechanisms with respect to their potential to increase the safety of a DNN and to use them as criteria for a benchmark.
This strategy represents a first draft to pave the way to an industrial consensus.
In future work, we will further evaluate the potential of the presented safety mechanisms and establish a concrete link to the DNN insufficiencies. Furthermore, we will work on concrete safety and data metrics to develop statistically relevant and valid evaluation metrics for the measurement of DNN safety.

\section*{Acknowledgement}
The research leading to these results is funded by the Federal Ministry for Economic Affairs and Energy within the project ``AI-Safeguarding - Methods and measures for safeguarding AI-based perception functions for automated driving''. The authors would like to thank the consortium for the successful cooperation. Special thanks go to Horst Michael Gro{\ss} from Ilmenau University of Technology, Neuroinformatics and Cognitive Robotics Lab, as well as Maram Akila, Christian Hellert, Sebastian Sudholt, Stephanie Abrecht, Loren Schwarz and Lydia Gauerhof for the fruitful discussions.

%
%
%
\bibliographystyle{splncs04}
\bibliography{egbib.bib}
%





\end{document}